\documentclass{article} 
\usepackage{iclr2021_conference,times}

\usepackage{amsmath,amsfonts,bm}

\def\eqref#1{equation~\ref{#1}}

\def\1{\bm{1}}

\DeclareMathAlphabet{\mathsfit}{\encodingdefault}{\sfdefault}{m}{sl}
\SetMathAlphabet{\mathsfit}{bold}{\encodingdefault}{\sfdefault}{bx}{n}

\usepackage{hyperref}
\usepackage{url}
\usepackage{svg}
\usepackage{float}

\title{On the Benefits of Early Fusion in Multimodal Representation Learning}

\author{George Barnum\thanks{Equal contributors.}, Sabera Talukder\footnotemark[1] \& Yisong Yue  \\
Department of Computation and Neural Systems, Neurobiology, Computing and Mathematical Sciences\\
California Institute of Technology \\
Pasadena, CA 91125, USA\\
\texttt{\{gbarnum,sabera,yyue\}@caltech.edu} \\
}

\iclrfinalcopy 
\begin{document}

\maketitle

\begin{abstract}
Intelligently reasoning about the world often requires integrating data from multiple modalities, as any individual modality may contain unreliable or incomplete information. Prior work in multimodal learning fuses input modalities only after significant independent processing. On the other hand, the brain performs multimodal processing almost immediately. This divide between conventional multimodal learning and neuroscience suggests that a detailed study of early multimodal fusion could improve artificial multimodal representations. To facilitate the study of early multimodal fusion, we create a convolutional LSTM network architecture that simultaneously processes both audio and visual inputs, and allows us to select the layer at which audio and visual information combines. Our results demonstrate that immediate fusion of audio and visual inputs in the initial C-LSTM layer results in higher performing networks that are more robust to the addition of white noise in both audio and visual inputs.

\end{abstract}



\section{Introduction}
Multimodal learning is important for many tasks, including audio visual speech recognition \citep{yu2020audio, zhou2019modality, su2017multimodal}, emotion recognition \citep{park2020k, cao2014crema}, multimedia event detection \citep{song2019review}, depth-based object detection \citep{wang2015large,wang2015mmss}, urban dynamics modeling \citep{zhang2017regions}, image-sentence matching \citep{liu2019modality}, and biometric recognition \citep{song2019review}. In many cases, an individual modality does not contain sufficient information to classify the scene. Therefore, utilizing multiple modalities is crucial, particularly in complex tasks or domains prone to noisy data.

One important design decision in multimodal  learning is how to best combine, or fuse, the different input modalities \citep{baltruvsaitis2018multimodal,li2018survey}.  
Prior work on multimodal learning has largely relied on extensive unimodal featurization and other preprocessing before fusing the  different modalities \citep{katsaggelos2015audiovisual,atrey2010multimodal}.  On the other hand, it is known that biological neural networks engage in  multimodal fusion in the very early layers of sensory processing pathways \citep{schroeder_multisensory_2005,budinger_multisensory_2006}.  This divide between conventional multimodal learning and neuroscience suggests that a detailed study of early multimodal fusion could yield insights for improving multimodal representation learning.

In this paper, we conduct a detailed study on the benefits of early fusion in multimodal representation learning, focusing on audio and visual modalities as they are the most related to human sensory processing.  To facilitate this study, we design a convolutional LSTM (C-LSTM) architecture that enables audio and visual input fusion at various layers in the architecture.  We find that fusion in the initial layer outperforms fusion before the fully connected (FC) layer, and that early fusion enables robust performance over a range of audio and visual signal to noise ratios (SNRs). We further study the interaction effects of fusion with noisy inputs both in one and two modalities. 
These results shed new light on the power of immediate fusion as a means to improve model performance in the presence of noise. Integrating multimodal inputs as soon as possible can be generalized to other multimodal domains, such as audio-visual speech recognition and emotion recognition to increase their performance and representational power.

\section{Related Work}

\textbf{Multimodal Representation Learning.}
\citet{baltruvsaitis2018multimodal} thoughtfully broke down the main problems within multimodal machine learning into five categories: representation, translation, alignment, fusion, and co-learning. Another helpful analysis of multimodal interactions revealed best practices while building mutlimodal systems: users tend to intermix unimodal and multimodal interactions, multimodal systems' main advantage occurs in decreasing errors or increasing flexibility \citep{turk2014multimodal}. However, in the same paper \citet{turk2014multimodal} did concede that multimodal integration,  also referred to as the fusion engine, is the key technical challenge for multimodal systems. 

When dealing with the problem of fusion, there are currently two common paradigms: early fusion \citep{atrey2010multimodal} and immediate fusion  \citep{katsaggelos2015audiovisual}. In early fusion, audio and visual modalities are first featurized before being passed to two independent modeling process units that do not differentiate between features from different modalities \citep{katsaggelos2015audiovisual}. On the other hand, immediate fusion is when the audio and visual modalities are first featurized and then sent to a join modeling process unit \citep{katsaggelos2015audiovisual}. This unfortunate terminology does not take into account the possibility of fusing the inputs before any substantial featurization, which does occur in biological neural networks.

A closely related area is multi-view learning \citep{li2018survey}.  While the two areas share significant overlap, multi-view learning places emphasis on having different views from the same input modality. A typical example is capturing the same scene from two viewing angles (where both views use the visual modality).

\textbf{Connections to Neuroscience.}
In the brain, multisensory integration was traditionally believed to occur only after single modality inputs underwent extensive processing in unisensory regions \citep{schroeder_multisensory_2005}. However, we now know that in many species, including humans, that multisensory convergence occurs much earlier in low level coritical structures \citep{schroeder_multisensory_2005}. In fact, primary sensory cortices may not be unimodal at all \citep{budinger_multisensory_2006}. This may in part be because of individual neuron's abilities to be modulated by multiple modalities \citep{meredith2009subthreshold}. In a striking discovery, \citet{allman2007multisensory} found that 16\% of visual neurons in the posterolateral lateral suprasylvian that were previously believed to be only visually responsive were significantly facilitated by auditory stimuli. This philosophical departure from individual modality processing towards early multimodal convergence in neuroscience lays a promising groundwork for high-impact explorations in multimodal machine learning.



\section{Multimodal Convolutional LSTM Model}
\label{sec:model}




Artificial neural networks often introduce inductive biases based on the structure of the input modality, such as convolution or recurrence in the case of visual or audio inputs \citep{44e2afaa580a48bc8b13633b22ff10b4,7472917, 7472176}. Since these biases are usually modality specific, approaches in multimodal domains frequently involve a degree of independent modality processing with a corresponding inductive bias. On the other hand, biological neural networks perform some multimodal processing almost immediately.\citep{schroeder_multisensory_2005,budinger_multisensory_2006}. 

\begin{figure}[t]
  \begin{minipage}{0.64\textwidth}
  \centering
  \includegraphics[width=1\textwidth]{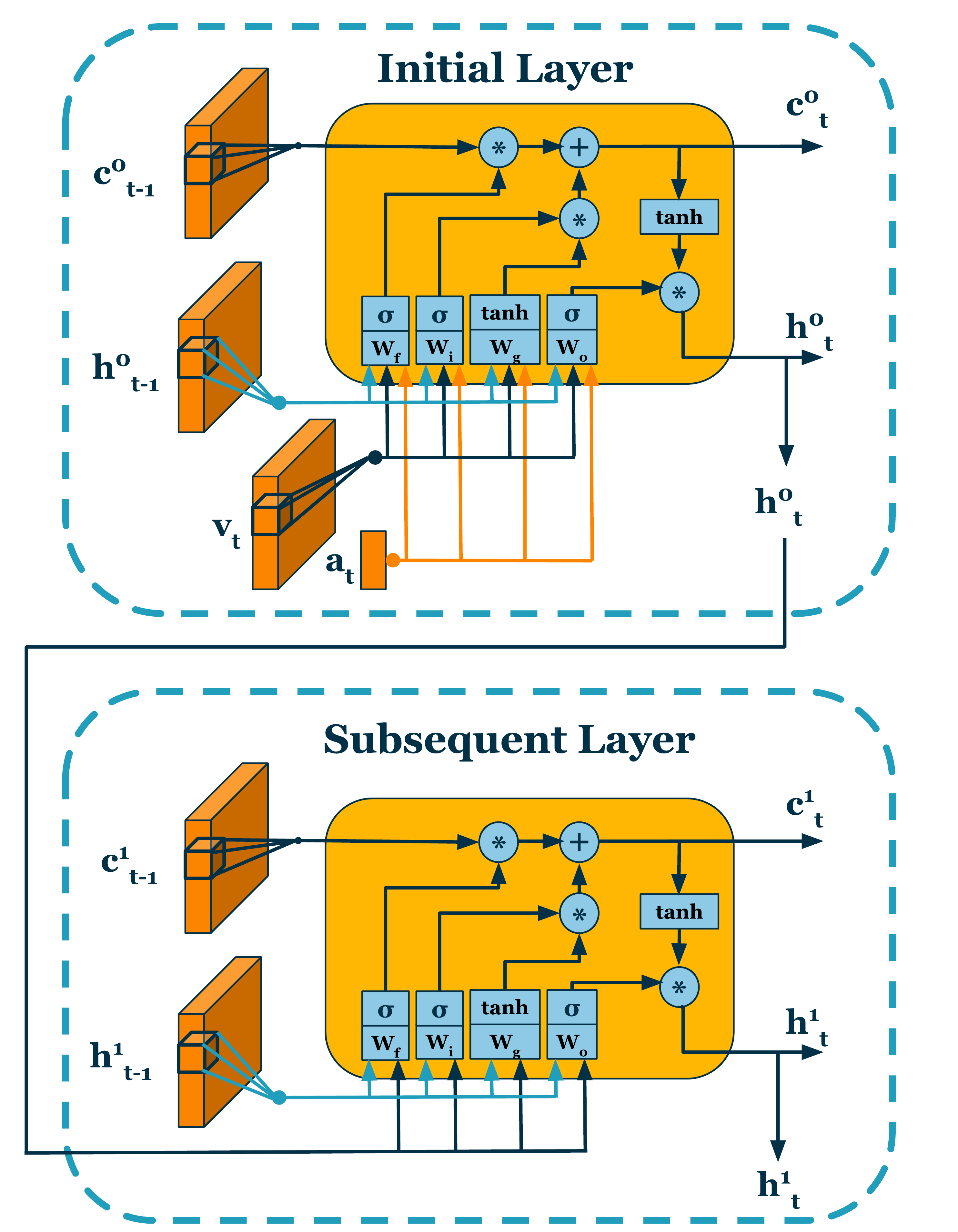}
\end{minipage}
  \begin{minipage}{0.35\textwidth}
  Initial LSTM Gate Values:
  \begin{align*}
f_t &= \sigma(W_{f}[\mathbf{h}_{t-1},\mathbf{x}_t]+ b_f)\\
i_t &= \sigma(W_{i}[\mathbf{h}_{t-1},\mathbf{x}_t]+ b_i)\\
g_t &=\tanh(W_{g}[\mathbf{h}_{t-1},\mathbf{x}_t] + b_g)\\
o_t &= \sigma(W_{o}[\mathbf{h}_{t-1},\mathbf{x}_t] + b_o)\\
\end{align*}
Cell State Update:
\begin{align*}
c_t = f_tc_{t-1}+i_tg_t\\
\end{align*}
Hidden State Update:
\begin{align*}
h_t = o_t\tanh(c_t)\\
\end{align*}
Subsequent LSTM Gate Values:
\begin{align*}
f_t &= \sigma(W_{f}[\mathbf{h}_{t-1},\mathbf{x}_t]+ b_f)\\
i_t &= \sigma(W_{i}[\mathbf{h}_{t-1},\mathbf{x}_t]+ b_i)\\
g_t &=\tanh(W_{g}[\mathbf{h}_{t-1},\mathbf{x}_t] + b_g)\\
o_t &= \sigma(W_{o}[\mathbf{h}_{t-1},\mathbf{x}_t] + b_o)\\
\end{align*}
  \end{minipage}

  \caption{
  The first two layers of the multimodal convolutional long-short term memory network, and the equations used to compute the gate and update values. $f_t$ is the forget gate, $i_t$ is the input gate, $g_t$ is the cell gate, and $o_t$ is the output gate. $W$'s are the corresponding weight matrices, and $b$'s the corresponding bias values. $\sigma()$ is the sigmoid function. $\tanh()$ is the hyperbolic tangent function.}
  \label{fig:clstm}
\end{figure}

In order to maintain the advantages of these modality specific inductive biases while also allowing for the immediate fusion of audio and visual inputs, we created a  multimodal convolutional long short term memory network that generates fused audio-visual representations with appropriate inductive biases.
Our convolutional long short term memory, or C-LSTM, architecture combines the convolutional properties found in traditional convolutional neural networks, and traditional long short term memory networks. At each point of convolution, the first layer takes:
\begin{itemize}
  \item The section of the input image to be multiplied by our convolutional kernel, denoted as $\mathbf{v}$.
  
  \item The section of the hidden state to be multiplied by our convolutional kernel, denoted as $\mathbf{h}_{t-1}$. It is initialized to zeros for the first time step in the first layer.
  
  \item The spectrogram value of the audio input, at a given time step, denoted as $\mathbf{a}_t$.
\end{itemize}
Our model then computes the LSTM gate values using: $\mathbf{v}$, $\mathbf{h}_{t-1}$, and $\mathbf{a}_t$ via the equations in Figure \ref{fig:clstm}, under the \textit{Initial LSTM Gate Values} section. 
The initial C-LSTM layer produces a 
single multimodal tensor that combines information from the audio and visual inputs. 
As in the standard LSTM architecture, the hidden state, $\mathbf{h}_t$, of the previous layer is used as the input, $\mathbf{x}_t$, of the current layer.
Therefore, in our subsequent LSTM layers the gate values at each location are computed from the section of the 
combined multimodal input,
$\mathbf{x}_t$, to be multiplied by our convolutional kernel. Equations found in Figure \ref{fig:clstm}, under the \textit{Subsequent LSTM Gate Values} section.  
By applying the LSTM operations at each location of a convolution, this architecture allows the LSTM cells to respond to the spatial information from the visual domain as well as the temporal information of the audio domain. This architecture enables us to study the mixing of signals at the initial, second, and fully connected layer while maintaining the same inductive biases that are beneficial in processing image and sequence data.



\textbf{Varying the Fusion Level.}
The full C-LSTM approach described above performs early fusion, i.e., mixing the modalities starting in the very first layer.  However, our framework can be easily modified to only allow for fusion in later layers, or to mask out one modality all together -- in particular by forcing certain weights to be zero.  As such, we can use the C-LSTM architecture to conduct a controlled inquiry into our research question.





\section{Experiments}

\subsection{Dataset}

We constructed a multimodal dataset based on the well-known MNIST dataset \citep{lecun2010mnist} and the Free Spoken Digit dataset \citep{zohar_jackson_2018_1342401}.
We selected these datasets because of their tractability. This allows us to combine them to create a multimodal task which we know will be solvable. This allows us to artificially manipulate the difficulty of the task via the addition of noise. We prefer a regime in which we can break the system to understand its principles of fusion and representation.

\begin{figure}[t]
  \centering
  \includegraphics[width=1\textwidth]{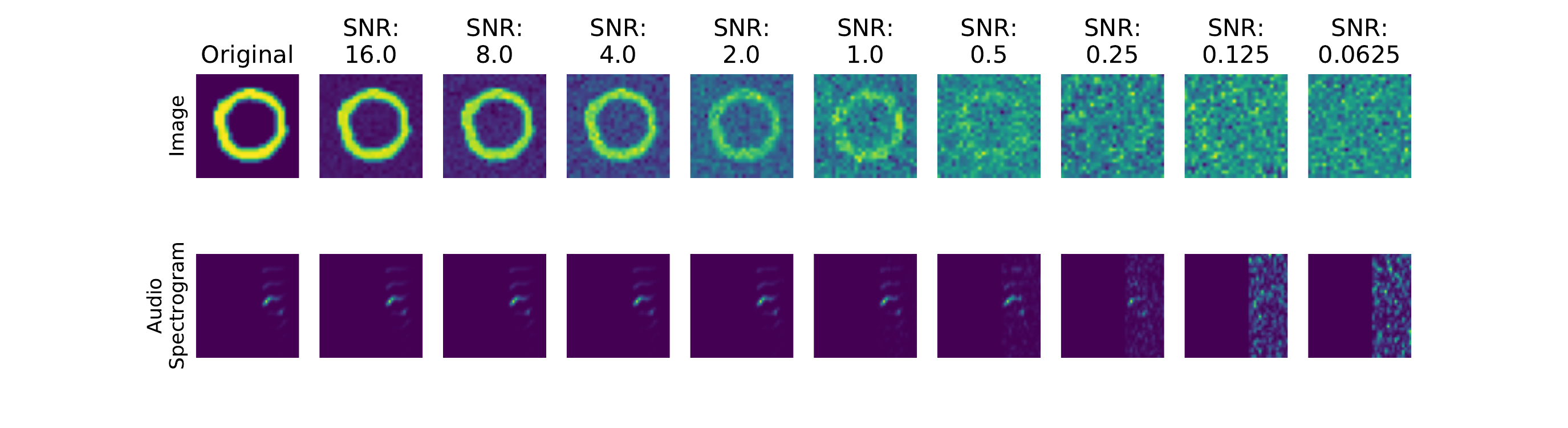}
  \vspace{-0.4in}
  \caption{A single input example where white noise has been applied at various SNR values. Audio spectrograms have been truncated to the first 42 frequency bins for convenience, with the full spectrograms available in Figure \ref{fig:data_full}}.
  \label{fig:data}
\end{figure}

For each of these datasets, we first combined all the data, testing and training, into one dataset, then split each of the two datasets into training, testing, and validation sets with an $8:1:1$ ratio. Next, for each of the training, testing and validation splits, we created a dataset containing all image and audio pairs with the same label. 
We then augmented the data by adding white noise to the image and audio data such that the signal to noise ratio could be chosen, as shown in Figure \ref{fig:data} and Appendix \ref{apend:data}. This allows us to explore our model's response to degradation in either the image or audio input, while utilizing noise techniques commonly applied to inputs \citep{borji2020whitenoise}.
We then front padded each audio input with zeros such that all the audio examples are of equal length. Finally, we took the spectrogram of each audio input, with 400 samples, 201 frequency bins, and a stride of 200 samples.

\subsection{Training \& Modeling Details}

Using our C-LSTM architecture described in Section \ref{sec:model}, we constructed multiple different models in order to study the benefits of multimodal fusion. These models can be viewed as ablations of the full model (i.e., fixing certain weights to zero).
\begin{itemize}
    \item The full C-LSTM model that allows for fusing in the early layers (akin to how biological networks fuse sensory processing in early layers).
    \item Restricting fusion to the intermediate convolutional layers.
    \item Restricting fusion to the fully connected layers (akin to prior work that performed unimodal featurization prior to fusion).
    \item Only processing visual or audio input.
\end{itemize}

Detailed model architectures are provided in \ref{apend:details}.



We trained all models using the Adam optimizer \citep{kingma2014adam} with  a learning rate of 0.001. Each model is trained on 87516 examples randomly selected from our multimodal dataset (which in total contains 11202076 training data points, 174490 validation data points, 175389 test data points). The models can be trained on any combination of audio and visual SNRs.

\subsection{Fusion Comparison}


\begin{figure}[t]
  \centering
  \includegraphics[width=1\textwidth]{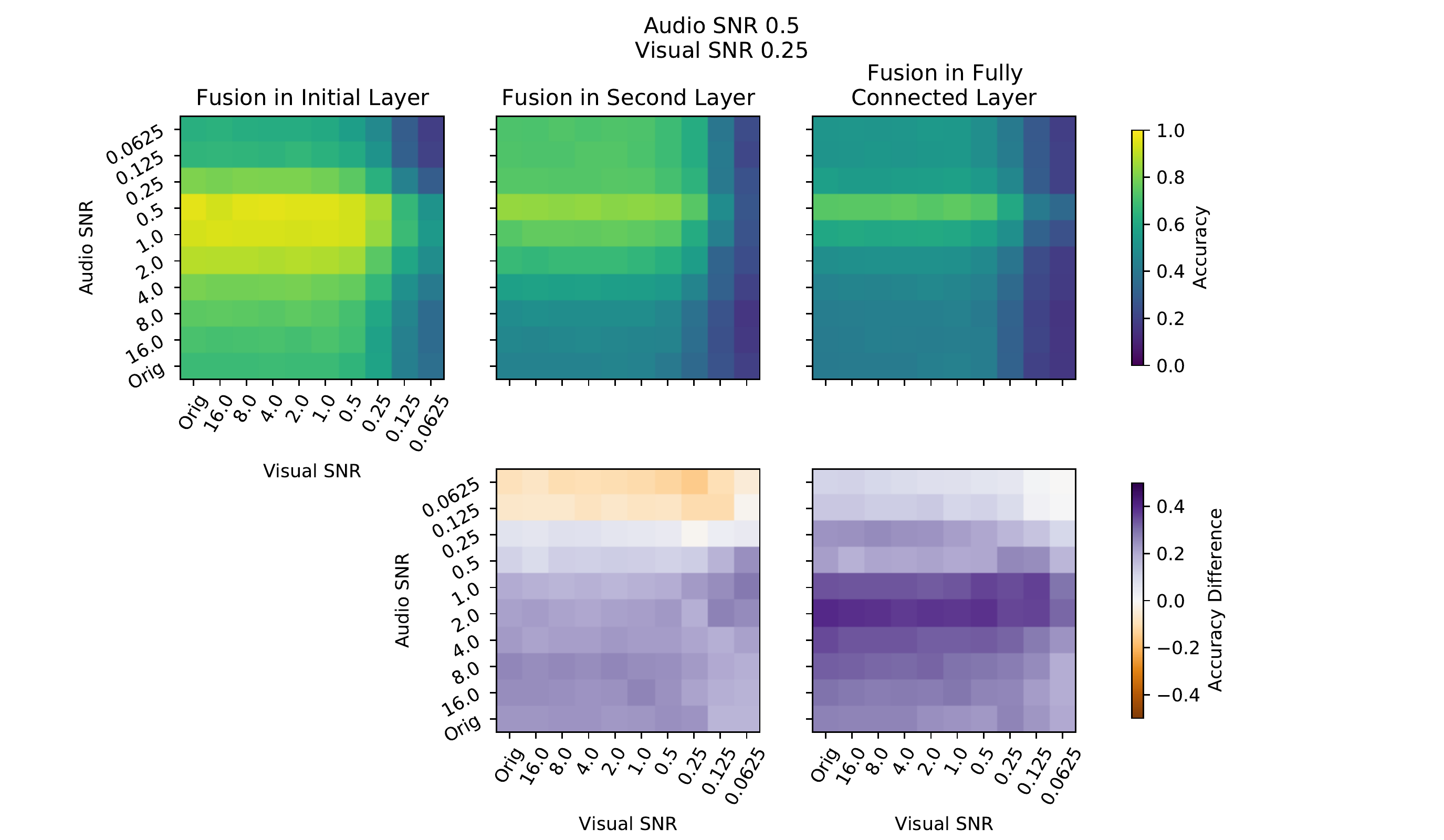}
  \vspace{-0.2in}
  \caption{Comparing the performance of initial layer fusion, second layer fusion, and FC fusion models. The first row shows raw test accuracy at various signal to noise ratios. The second row shows the difference in accuracy between the late fusion models and the immediate fusion model at corresponding signal to noise levels. All models were trained with an audio SNR of 0.5 and a visual SNR of 0.25. Orig signifies the original audio or visual input.}
  \label{fig:fusion}
\end{figure}



To examine the value of early compared to late multimodal fusion, the immediate fusion model, the second layer fusion model, and the fully connected layer fusion model were trained with the same signal to noise ratios of training data. 
We then tested the accuracy of each of these models for a range of values of the signal to noise ratios of both the audio and visual inputs. Then we compare the accuracy of the immediate fusion model to the accuracy of each of the late fusion models.

As seen in Figure \ref{fig:fusion}, the immediate fusion model is more accurate not only for the signal to noise ratio that the models were trained at, but also for the majority of the other signal to noise ratios. In particular, the initial layer fusion model always outperforms or equally performs to the fully connected fusion model. Furthermore, the initial layer fusion model outperforms the second layer fusion model, except for when the audio input is degraded well beyond the audio training signal to noise ratio. 
 

Additionally, initial layer fusion appears to allow the model to be much more robust to increases in the signal to noise ratio beyond the training values, especially in the case of increased audio SNR. The main characteristic of the SNR values in which immediate fusion does not outperform late fusion is low audio SNR and relatively high visual SNR, and this only occurs in the case where fusion occurs in the second layer. This suggests that immediate multimodal fusion encourages the multimodal model to use both input modalities.

While the specific areas and degree to which early fusion outperforms delayed fusion varies with the SNRs of the training data, the general trends are similar, and these fusion plots are representative of models trained at other audio and visual SNRs; see \ref{apend:fusion} for the same plot at other audio and visual training levels.

\subsection{Robustness to Noise}

\begin{figure}[t]
  \centering
  \includegraphics[width=1\textwidth]{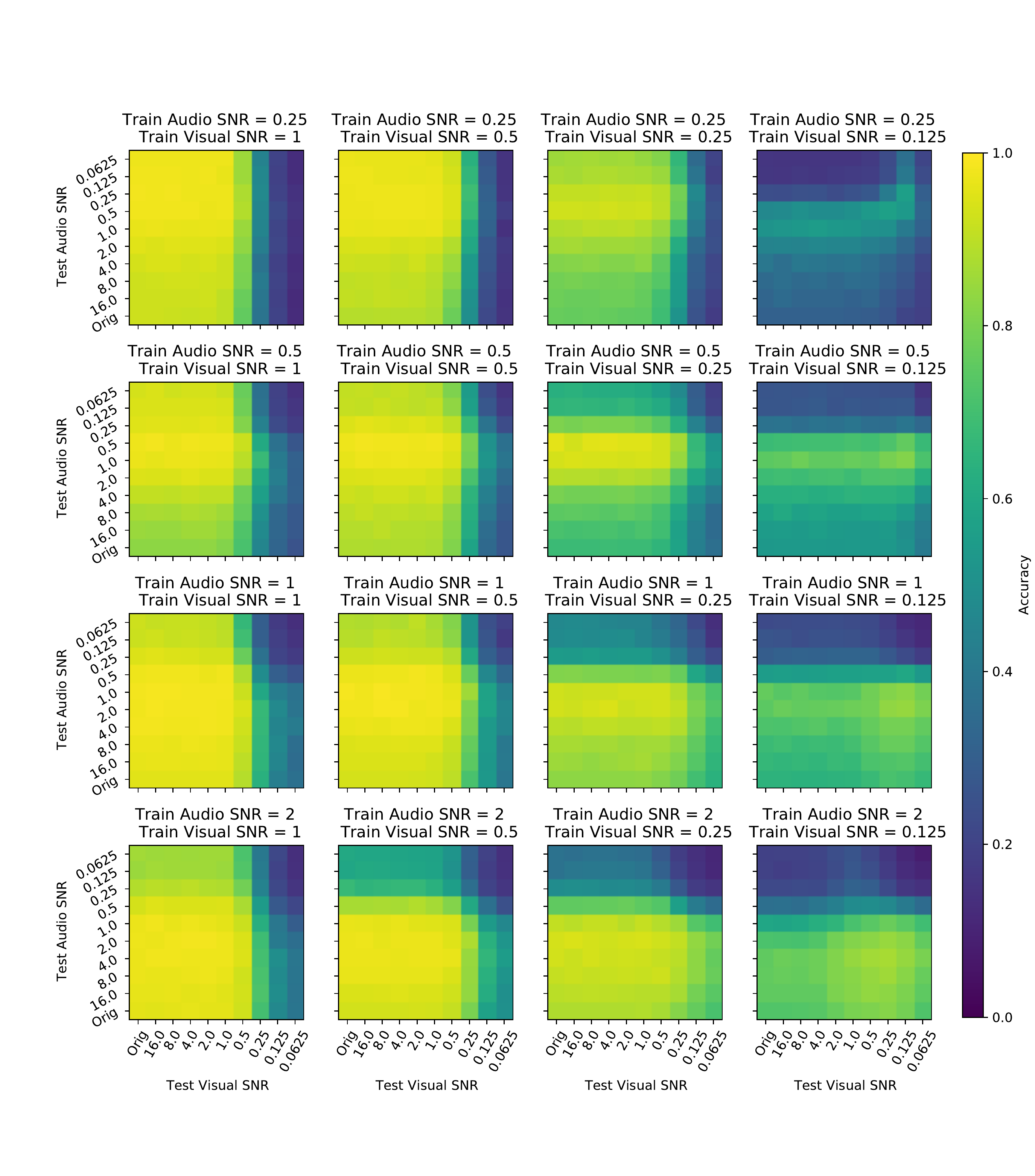}
  \vspace{-0.6in}
  \caption{Accuracy of multimodal models trained on data with a range of signal to noise ratios, each for a range of signal to noise ratios of the testing data. 
  }
  \label{fig:noise}
\end{figure}

An advantage of multimodal processing is the network's resilience to noise in the inputs. To examine our multimodal model's resilience to white noise we trained 16 models at distinct audio-visual SNR combinations and then tested each model in 100 different audio-visual SNR regimes. The results of these experiments can be seen in figure \ref{fig:noise}.

We see that in columns where the visual SNR is 1 and 0.5, there is strong visual dependence. However, MNIST is a limited dataset, where the original images are nicely positioned in the field of view and are relatively noise free. We believe that when adapting the principle of early multimodal fusion to real world datasets, one would not have as clearly defined visual or audio dependence. The principles that we elicit from performing this experiment with MNIST should guide us towards understanding fusion, while recognizing that the datasets and real life scenarios this will be adapted to will be much more complex and likely necessitate both modalities at a range of both audio and visual SNRs. 
The addition of noise to both the training and test data is designed to provide a setting that allows the exploration of the limits of this model while using a simple dataset.

In the the top left quadrant of the figure, we see that at higher visual SNRs (1, 0.5) and lower audio SNRs (0.5, 0.25) the model is mainly dependent on visual information. This is denoted by the test accuracy's consistency as the audio SNR varies from the original to 0.0625. As we move to the bottom right quadrant we see lower visual SNRs (0.25, 0.125) and higher audio SNRs (2, 1) accompanied by an increasing dependence on the audio information. This increasing dependence is indicated by the test accuracy's consistency as the visual SNR varies from original to 0.0625. In the bottom left quadrant, which corresponds to high audio and visual SNRs, the model performs well on all of the combinations of audio and visual SNRs that are larger in value than the audio-visual SNR that the model was trained on. The lower performance in the SNR ranges below what the model was trained in either the audio or visual situations is expected. Unsurprisingly, in the top right quadrant, which corresponds to low audio and visual SNRs, we see that the test accuracy of the model falls. In these low SNR regimes, the poor signal quality in both modalities results in poor performance outside of the audio-visual SNRs the model was tested on. These results mirror our expectations of how a multimodal model would behave both to various training and testing SNRs.



\subsection{Comparison to Unimodal Models}



\begin{figure}[ht]
  \centering
  \includegraphics[width=1\textwidth]{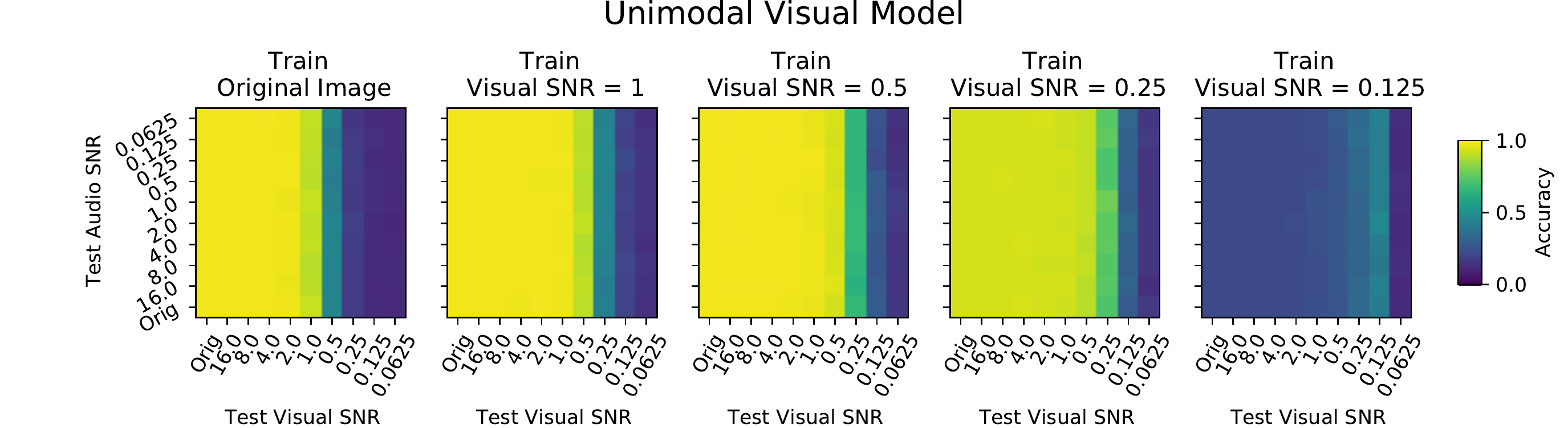}
  \vspace{-0.2in}
  \caption{Performance of the C-LSTM model with only visual input trained on the original data as well as at various visual signal to noise ratios.}
  \label{fig:audio_knockout}
\end{figure}
\begin{figure}[ht]
  \centering
  \includegraphics[width=1\textwidth]{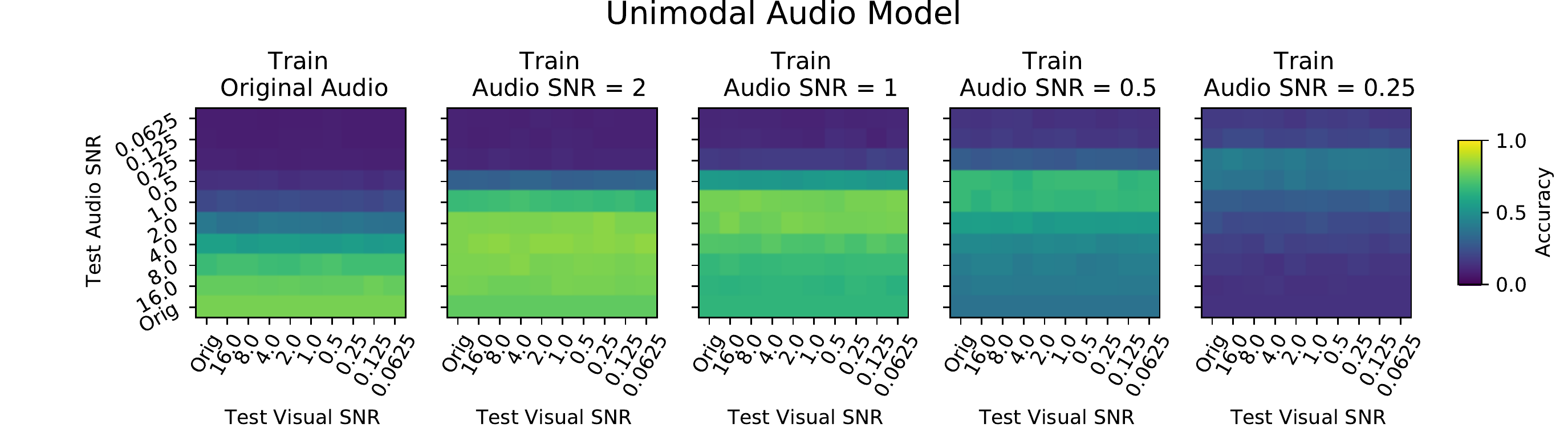}
  \vspace{-0.2in}
  \caption{
  Performance of the C-LSTM model with only audio input trained on the original data as well as at various audio signal to noise ratios. }
  \label{fig:visual_knockout}
\end{figure}

To verify that joint audio visual representations are a result of both modalities, we tested our multimodal model on unimodal inputs by setting one input to zero. This created unimodal visual models and unimodal audio models without changing the underlying architecture. For each of these unimodal models, we trained the model at four SNR values, and tested the accuracy across our previously selected set of signal to noise ratios. The accuracy of these unimodal models for the SNR values is displayed in figures \ref{fig:audio_knockout} and \ref{fig:visual_knockout}.

As expected, the unimodal models perform relatively well at the SNR that they are trained at or higher, although as the training SNR decreases, the models perform worse overall, and the performance on higher SNR data decreases. This effect is particularly noticeable in the audio only model, figure \ref{fig:visual_knockout}.

Additionally, these unimodal models prove that the C-LSTM architecture is at minimum capable of using information from either modality to perform the classification task. However, the difference in performance between the audio and visual unimodal models shows that the model is not equally sensitive to each modality or to noise in each modality. This discrepancy motivated us to choose an audio SNR of 0.5 and a visual SNR of 0.25 for further investigation of the model architecture, such that the contribution of each modality will be comparable and will allow us to investigate early fusion in a setting where multimodal fusion will be beneficial.


\subsection{Model Inspection}

\begin{figure}[ht]
  \centering
  \includegraphics[width=1\textwidth]{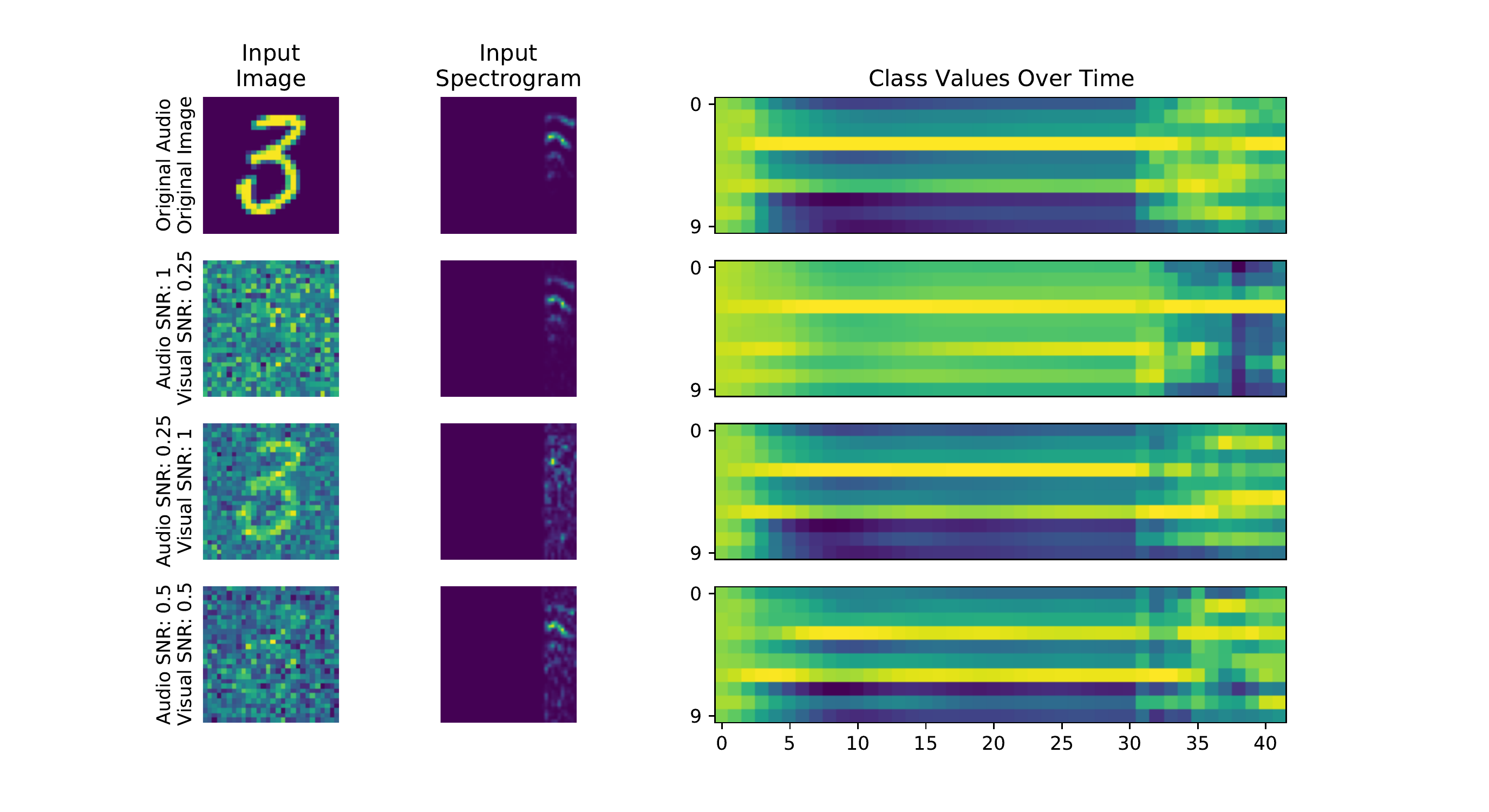}
  \vspace{-0.4in}
  \caption{The values of the final layer of the multimodal layer across timesteps of the C-LSTM for a representative example at various signal to noise ratios.}
  \label{fig:out}
\end{figure}

In order to examine the contribution of audio and visual inputs to the performance of our multimodal classifier, we considered the state of the network at intermediate timesteps in the recurrent processing of the audio inputs. In figure \ref{fig:out} we display the activations of the final layer of the network across timesteps for a single representative example at various signal to noise ratios. These final layer activations correspond to the classification of the input into each of the ten digit classes. We display the activations in response to four SNR scenarios: the original input, a scenario where the audio input has a higher SNR than the visual input, a scenario where the visual input has a higher SNR than the audio input, and a scenario where the visual input and the audio input have equal SNR. Additional examples are included in section \ref{apend:activation}.

This visualization demonstrates the value of the multimodal input in this network. Because the image information is available to the network for the entire length of the audio, the network initially responds to the image without audio input, because the audio is front zero-padded. As the network is evaluated along the time dimension of the audio input, information from the audio input becomes available to the network. Therefore, comparing the activations of the final layer can illuminate the contribution of each modality to the network.

In this example, the original input demonstrates the capability of the same model to correctly classify the digit in both the visual and audio regime, because both before and during the audio input, the correct class, 3, is assigned the maximum value. In the example where the audio input has an SNR of 1.0 and the visual input has an SNR of 0.25, as well as the example with audio SNR of 0.5 and visual SNR of 0.5, the contribution of the audio input becomes more evident, as in both of these scenarios, before the audio input is available the model shows more confusion, both with all the other classes, and in particular with class 6. In these examples, as the audio information becomes available to the network, the correct class comes to dominate the activations.


\section{Conclusion}

In this paper, we developed a C-LSTM architecture to investigate the effects of fusion depth on noise robustness. 
Motivated by the neuroscientific literature suggesting that sensory inputs are combined early in processing, we proposed that truly immediate fusion of modalities would provide benefits in neural networks as well.
Our immediate fusion model demonstrates robustness to changes in input noise and an improvement in accuracy relative to late fusion models with analogous architectures.
Future directions include investigating the effects of immediate multimodal fusion in deeper networks and on problems with more inherent difficulty, as well as the extension of immediate multimodal processing to other multimodal domains such as translation, alignment, and co-learning. However, we believe that the results demonstrated here show the importance of truly immediate fusion and could help with other domain specific multimodal learning tasks.






\bibliography{multimodal}

\begin{thebibliography}{26}
\providecommand{\natexlab}[1]{#1}
\providecommand{\url}[1]{\texttt{#1}}
\expandafter\ifx\csname urlstyle\endcsname\relax
  \providecommand{\doi}[1]{doi: #1}\else
  \providecommand{\doi}{doi: \begingroup \urlstyle{rm}\Url}\fi

\bibitem[Allman \& Meredith(2007)Allman and Meredith]{allman2007multisensory}
Brian~L Allman and M~Alex Meredith.
\newblock Multisensory processing in “unimodal” neurons: cross-modal
  subthreshold auditory effects in cat extrastriate visual cortex.
\newblock \emph{Journal of neurophysiology}, 98\penalty0 (1):\penalty0
  545--549, 2007.

\bibitem[Atrey et~al.(2010)Atrey, Hossain, El~Saddik, and
  Kankanhalli]{atrey2010multimodal}
Pradeep~K Atrey, M~Anwar Hossain, Abdulmotaleb El~Saddik, and Mohan~S
  Kankanhalli.
\newblock Multimodal fusion for multimedia analysis: a survey.
\newblock \emph{Multimedia systems}, 16\penalty0 (6):\penalty0 345--379, 2010.

\bibitem[Baltru{\v{s}}aitis et~al.(2018)Baltru{\v{s}}aitis, Ahuja, and
  Morency]{baltruvsaitis2018multimodal}
Tadas Baltru{\v{s}}aitis, Chaitanya Ahuja, and Louis-Philippe Morency.
\newblock Multimodal machine learning: A survey and taxonomy.
\newblock \emph{IEEE transactions on pattern analysis and machine
  intelligence}, 41\penalty0 (2):\penalty0 423--443, 2018.

\bibitem[Borji \& Lin(2020)Borji and Lin]{borji2020whitenoise}
Ali Borji and Sikun Lin.
\newblock White noise analysis of neural networks.
\newblock In \emph{International Conference on Learning Representations}, 2020.
\newblock URL \url{https://openreview.net/forum?id=H1ebhnEYDH}.

\bibitem[Budinger et~al.(2006)Budinger, Heil, Hess, and
  Scheich]{budinger_multisensory_2006}
E.~Budinger, P.~Heil, A.~Hess, and H.~Scheich.
\newblock Multisensory processing via early cortical stages: Connections of the
  primary auditory cortical field with other sensory systems.
\newblock \emph{Neuroscience}, 143\penalty0 (4):\penalty0 1065--1083, 2006.
\newblock ISSN 0306-4522.
\newblock \doi{10.1016/j.neuroscience.2006.08.035}.
\newblock URL
  \url{http://www.sciencedirect.com/science/article/pii/S0306452206011158}.

\bibitem[Cao et~al.(2014)Cao, Cooper, Keutmann, Gur, Nenkova, and
  Verma]{cao2014crema}
Houwei Cao, David~G Cooper, Michael~K Keutmann, Ruben~C Gur, Ani Nenkova, and
  Ragini Verma.
\newblock Crema-d: Crowd-sourced emotional multimodal actors dataset.
\newblock \emph{IEEE transactions on affective computing}, 5\penalty0
  (4):\penalty0 377--390, 2014.

\bibitem[Jackson et~al.(2018)Jackson, Souza, Flaks, Pan, Nicolas, and
  Thite]{zohar_jackson_2018_1342401}
Zohar Jackson, César Souza, Jason Flaks, Yuxin Pan, Hereman Nicolas, and
  Adhish Thite.
\newblock Jakobovski/free-spoken-digit-dataset: v1.0.8, August 2018.
\newblock URL \url{https://doi.org/10.5281/zenodo.1342401}.

\bibitem[Katsaggelos et~al.(2015)Katsaggelos, Bahaadini, and
  Molina]{katsaggelos2015audiovisual}
Aggelos~K Katsaggelos, Sara Bahaadini, and Rafael Molina.
\newblock Audiovisual fusion: Challenges and new approaches.
\newblock \emph{Proceedings of the IEEE}, 103\penalty0 (9):\penalty0
  1635--1653, 2015.

\bibitem[Kingma \& Ba(2014)Kingma and Ba]{kingma2014adam}
Diederik~P. Kingma and Jimmy Ba.
\newblock Adam: A method for stochastic optimization, 2014.

\bibitem[Lecun \& Bengio(1995)Lecun and
  Bengio]{44e2afaa580a48bc8b13633b22ff10b4}
Yann Lecun and Yoshua Bengio.
\newblock \emph{Convolutional networks for images, speech, and time-series}.
\newblock MIT Press, 1995.

\bibitem[LeCun et~al.(2010)LeCun, Cortes, and Burges]{lecun2010mnist}
Yann LeCun, Corinna Cortes, and CJ~Burges.
\newblock Mnist handwritten digit database.
\newblock \emph{ATT Labs [Online]. Available:
  http://yann.lecun.com/exdb/mnist}, 2, 2010.

\bibitem[Li et~al.(2018)Li, Yang, and Zhang]{li2018survey}
Yingming Li, Ming Yang, and Zhongfei Zhang.
\newblock A survey of multi-view representation learning.
\newblock \emph{IEEE transactions on knowledge and data engineering},
  31\penalty0 (10):\penalty0 1863--1883, 2018.

\bibitem[Liu et~al.(2019)Liu, Zhao, Wei, Zheng, and Yang]{liu2019modality}
Ruoyu Liu, Yao Zhao, Shikui Wei, Liang Zheng, and Yi~Yang.
\newblock Modality-invariant image-text embedding for image-sentence matching.
\newblock \emph{ACM Transactions on Multimedia Computing, Communications, and
  Applications (TOMM)}, 15\penalty0 (1):\penalty0 1--19, 2019.

\bibitem[Meredith \& Allman(2009)Meredith and Allman]{meredith2009subthreshold}
M~Alex Meredith and Brian~L Allman.
\newblock Subthreshold multisensory processing in cat auditory cortex.
\newblock \emph{Neuroreport}, 20\penalty0 (2):\penalty0 126, 2009.

\bibitem[{Parascandolo} et~al.(2016){Parascandolo}, {Huttunen}, and
  {Virtanen}]{7472917}
G.~{Parascandolo}, H.~{Huttunen}, and T.~{Virtanen}.
\newblock Recurrent neural networks for polyphonic sound event detection in
  real life recordings.
\newblock In \emph{2016 IEEE International Conference on Acoustics, Speech and
  Signal Processing (ICASSP)}, pp.\  6440--6444, 2016.

\bibitem[Park et~al.(2020)Park, Cha, Kang, Kim, Khandoker, Hadjileontiadis, Oh,
  Jeong, and Lee]{park2020k}
Cheul~Young Park, Narae Cha, Soowon Kang, Auk Kim, Ahsan~Habib Khandoker,
  Leontios Hadjileontiadis, Alice Oh, Yong Jeong, and Uichin Lee.
\newblock K-emocon, a multimodal sensor dataset for continuous emotion
  recognition in naturalistic conversations.
\newblock \emph{arXiv preprint arXiv:2005.04120}, 2020.

\bibitem[Schroeder \& Foxe(2005)Schroeder and
  Foxe]{schroeder_multisensory_2005}
Charles~E Schroeder and John Foxe.
\newblock Multisensory contributions to low-level, 'unisensory' processing.
\newblock \emph{Current Opinion in Neurobiology}, 15\penalty0 (4):\penalty0
  454--458, 2005.
\newblock ISSN 0959-4388.
\newblock \doi{10.1016/j.conb.2005.06.008}.
\newblock URL
  \url{http://www.sciencedirect.com/science/article/pii/S0959438805000991}.

\bibitem[Song et~al.(2019)Song, Chen, Wang, Chen, Tian, and
  Tang]{song2019review}
Xiaoyu Song, Hong Chen, Qing Wang, Yunqiang Chen, Mengxiao Tian, and Hui Tang.
\newblock A review of audio-visual fusion with machine learning.
\newblock In \emph{Journal of Physics: Conference Series}, volume 1237, pp.\
  022144. IOP Publishing, 2019.

\bibitem[Su et~al.(2017)Su, Wang, and Liu]{su2017multimodal}
Rongfeng Su, Lan Wang, and Xunying Liu.
\newblock Multimodal learning using 3d audio-visual data for audio-visual
  speech recognition.
\newblock In \emph{2017 International Conference on Asian Language Processing
  (IALP)}, pp.\  40--43. IEEE, 2017.

\bibitem[Turk(2014)]{turk2014multimodal}
Matthew Turk.
\newblock Multimodal interaction: A review.
\newblock \emph{Pattern Recognition Letters}, 36:\penalty0 189--195, 2014.

\bibitem[Wang et~al.(2015{\natexlab{a}})Wang, Cai, Lu, and Cham]{wang2015mmss}
Anran Wang, Jianfei Cai, Jiwen Lu, and Tat-Jen Cham.
\newblock Mmss: Multi-modal sharable and specific feature learning for rgb-d
  object recognition.
\newblock In \emph{Proceedings of the IEEE international conference on computer
  vision}, pp.\  1125--1133, 2015{\natexlab{a}}.

\bibitem[Wang et~al.(2015{\natexlab{b}})Wang, Lu, Cai, Cham, and
  Wang]{wang2015large}
Anran Wang, Jiwen Lu, Jianfei Cai, Tat-Jen Cham, and Gang Wang.
\newblock Large-margin multi-modal deep learning for rgb-d object recognition.
\newblock \emph{IEEE Transactions on Multimedia}, 17\penalty0 (11):\penalty0
  1887--1898, 2015{\natexlab{b}}.

\bibitem[{Wang} et~al.(2016){Wang}, {Neves}, and {Metze}]{7472176}
Y.~{Wang}, L.~{Neves}, and F.~{Metze}.
\newblock Audio-based multimedia event detection using deep recurrent neural
  networks.
\newblock In \emph{2016 IEEE International Conference on Acoustics, Speech and
  Signal Processing (ICASSP)}, pp.\  2742--2746, 2016.

\bibitem[Yu et~al.(2020)Yu, Zhang, Wu, Ghorbani, Wu, Kang, Liu, Liu, Meng, and
  Yu]{yu2020audio}
Jianwei Yu, Shi-Xiong Zhang, Jian Wu, Shahram Ghorbani, Bo~Wu, Shiyin Kang,
  Shansong Liu, Xunying Liu, Helen Meng, and Dong Yu.
\newblock Audio-visual recognition of overlapped speech for the lrs2 dataset.
\newblock In \emph{ICASSP 2020-2020 IEEE International Conference on Acoustics,
  Speech and Signal Processing (ICASSP)}, pp.\  6984--6988. IEEE, 2020.

\bibitem[Zhang et~al.(2017)Zhang, Zhang, Yuan, Peng, Zheng, Hanratty, Wang, and
  Han]{zhang2017regions}
Chao Zhang, Keyang Zhang, Quan Yuan, Haoruo Peng, Yu~Zheng, Tim Hanratty,
  Shaowen Wang, and Jiawei Han.
\newblock Regions, periods, activities: Uncovering urban dynamics via
  cross-modal representation learning.
\newblock In \emph{Proceedings of the 26th International Conference on World
  Wide Web}, pp.\  361--370, 2017.

\bibitem[Zhou et~al.(2019)Zhou, Yang, Chen, Wang, and Jia]{zhou2019modality}
Pan Zhou, Wenwen Yang, Wei Chen, Yanfeng Wang, and Jia Jia.
\newblock Modality attention for end-to-end audio-visual speech recognition.
\newblock In \emph{ICASSP 2019-2019 IEEE International Conference on Acoustics,
  Speech and Signal Processing (ICASSP)}, pp.\  6565--6569. IEEE, 2019.

\end{thebibliography}
\bibliographystyle{iclr2021_conference}

\appendix
\section{Appendix}

\subsection{Input data}
\label{apend:data}

\begin{figure}[H]
  \centering
  \includegraphics[width=1\textwidth]{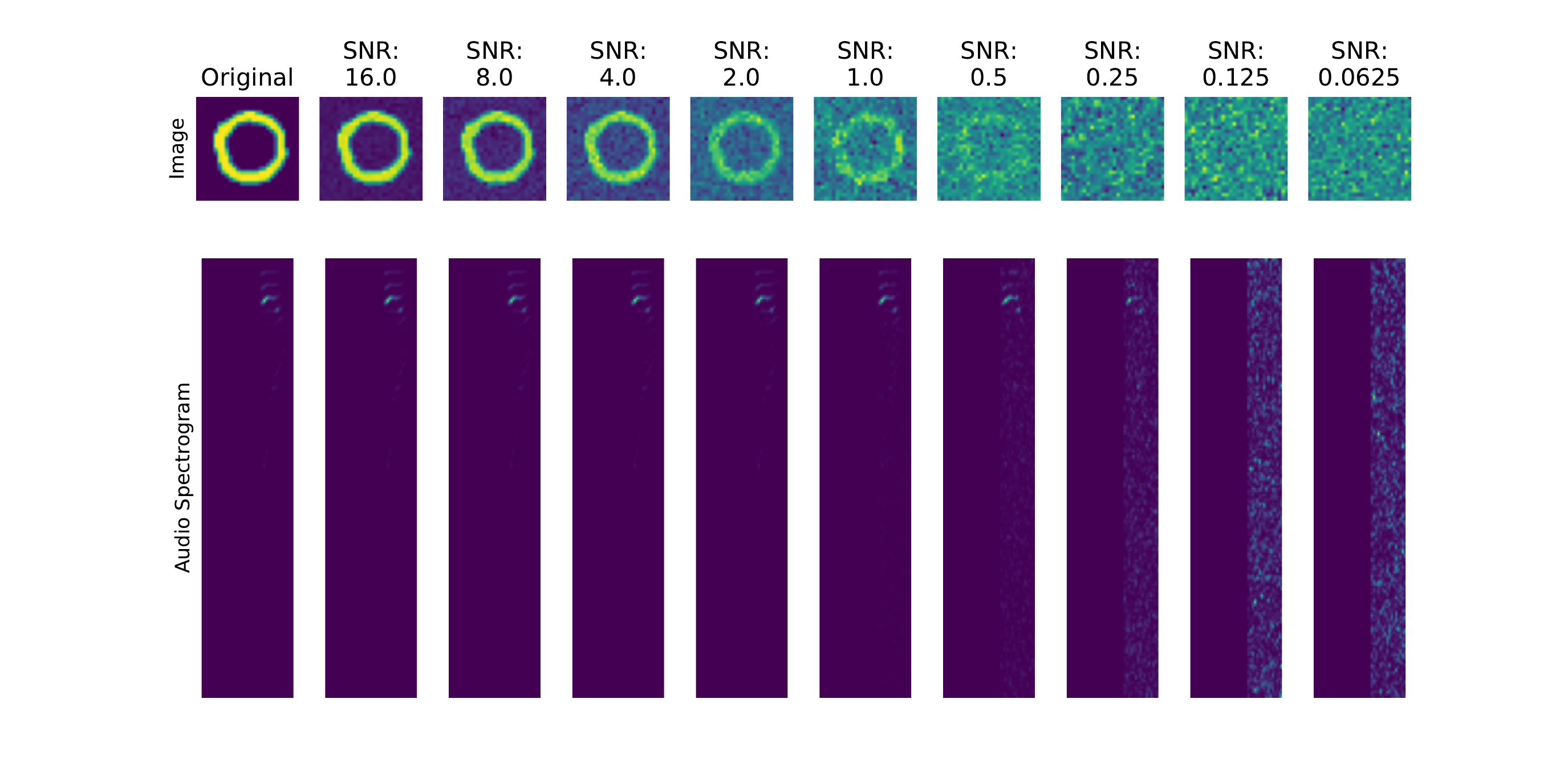}
  \vspace{-0.3in}
  \caption{The same input example where white noise has been applied at various SNR values, with the full audio spectrogram.}
  \label{fig:data_full}
\end{figure}

\subsection{Model Details}
\label{apend:details}

For the full model, we used: an initial merge layer (64 units and $3\times3$ kernels, and a $2\times2$ max pool layer), a second multimodal C-LSTM layer (64 units and $3\times3$ kernels, and a $2\times2$ max pool layer), a dense layer (128 units and ReLU activation), and then a final dense output layer (10 units). 

The second layer fusion model consists of a separate convolutional layer with 64 units and $3\times3$ kernels and an LSTM layer with 64 units. These layers feed into a C-LSTM layer, with 64 units and $3\times3$ kernels, a $2\times2$ max pool layer, a dense layer with 128 units and ReLU activation, and finally a final dense output layer with 10 units.

The fully connected layer fusion model consists of separate processing streams for the visual and audio data. The visual stream consists of two convolutional layers with 64 units and $3\times3$ kernels, while the audio stream consisting of two LSTM layers with 64 units. The output of the convolutional layers and the last timestep of the output of the LSTM layers are concatenated and fed into a dense layer with 128 units and ReLU activation, then a final dense output layer with 10 units.  

\subsection{Late Fusion}
\label{apend:fusion}

\begin{figure}[H]
  \centering
  \includegraphics[width=1\textwidth]{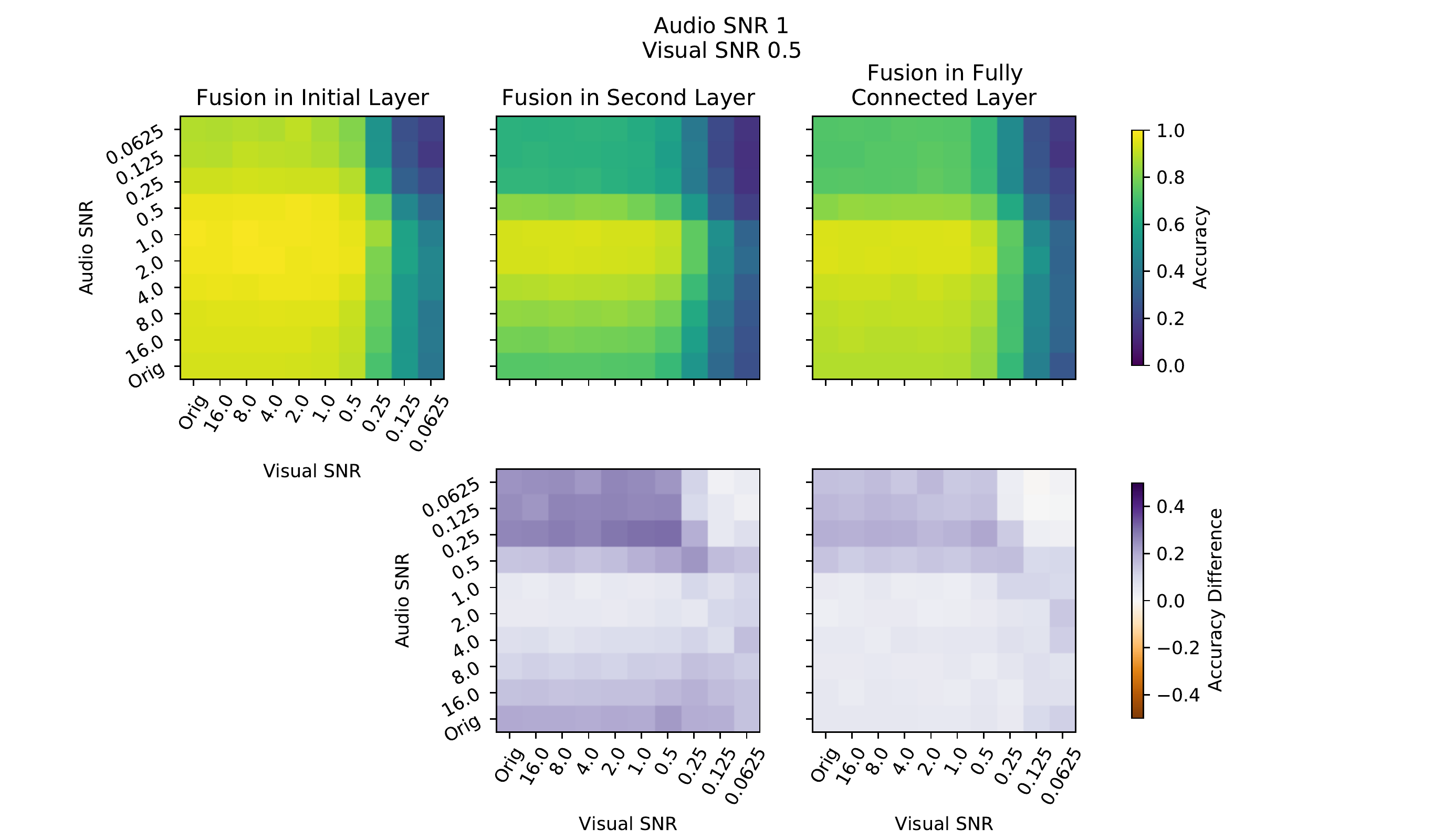}
  \vspace{-0.2in}
  \caption{Late fusion model where audio SNR = 1, and visual SNR = 0.5. Fusion in the initial layer outperforms the fusion in the fully connected layer.}
  \label{fig:late_fusion_a1_v0.5}
\end{figure}

\begin{figure}[h!]
  \centering
  \includegraphics[width=1\textwidth]{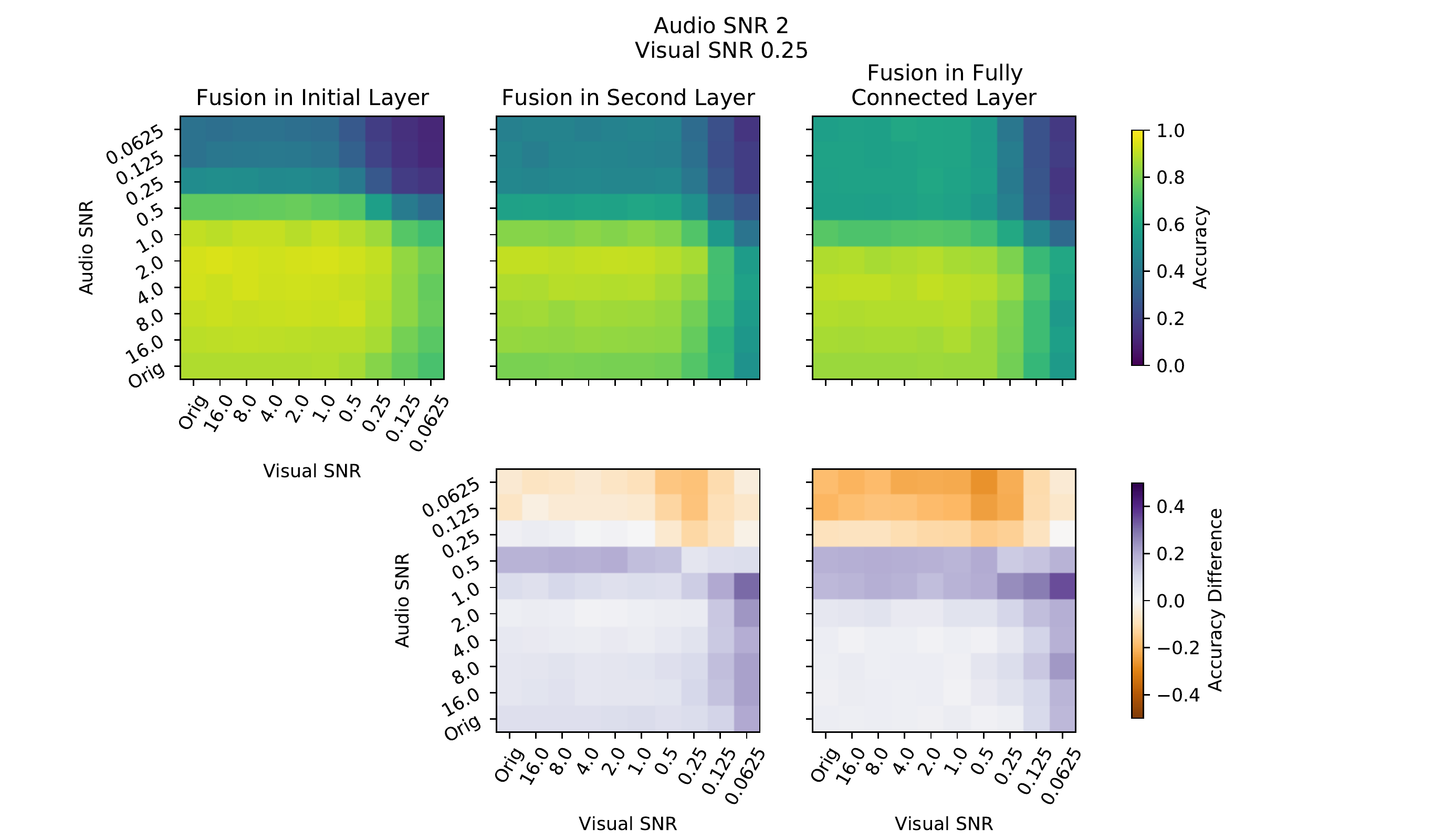}
  \vspace{-0.2in}
  \caption{Late fusion model where audio SNR = 2, and visual SNR = 0.25. Fusion in the initial layer outperforms the fusion in the fully connected layer for audio SNR values greater than 1.}
  \label{fig:late_fusion_a2_v0.25}
\end{figure}

\begin{figure}[h!]
  \centering
  \includegraphics[width=1\textwidth]{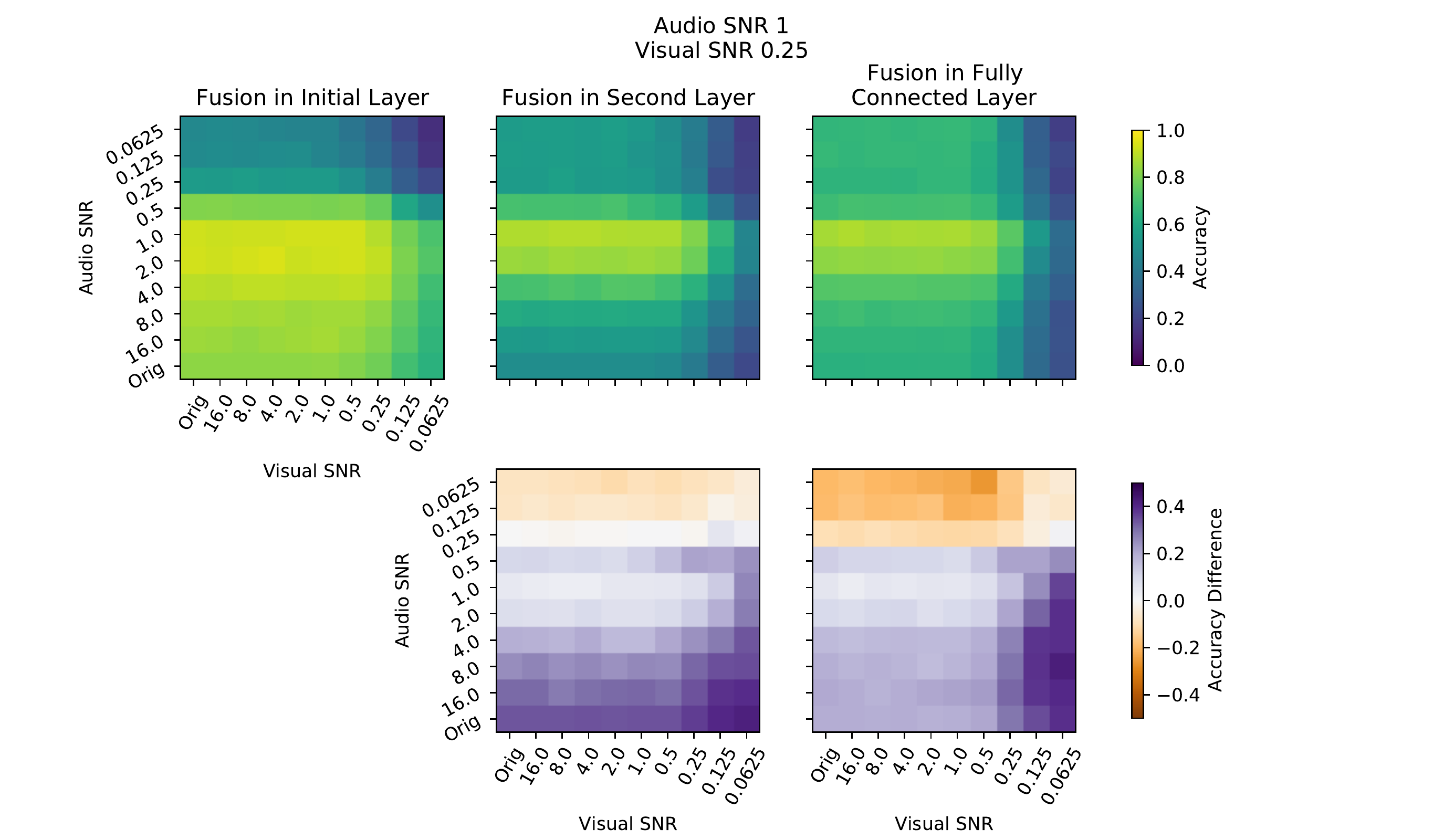}
  \vspace{-0.2in}
  \caption{Late fusion model where audio SNR = 1.0, and visual SNR = 0.25. Fusion in the initial layer outperforms the fusion in the fully connected layer for audio SNR values greater than 1.}
  \label{fig:late_fusion_a1_v0.25}
\end{figure}


\begin{figure}[h!]
  \centering
  \includegraphics[width=1\textwidth]{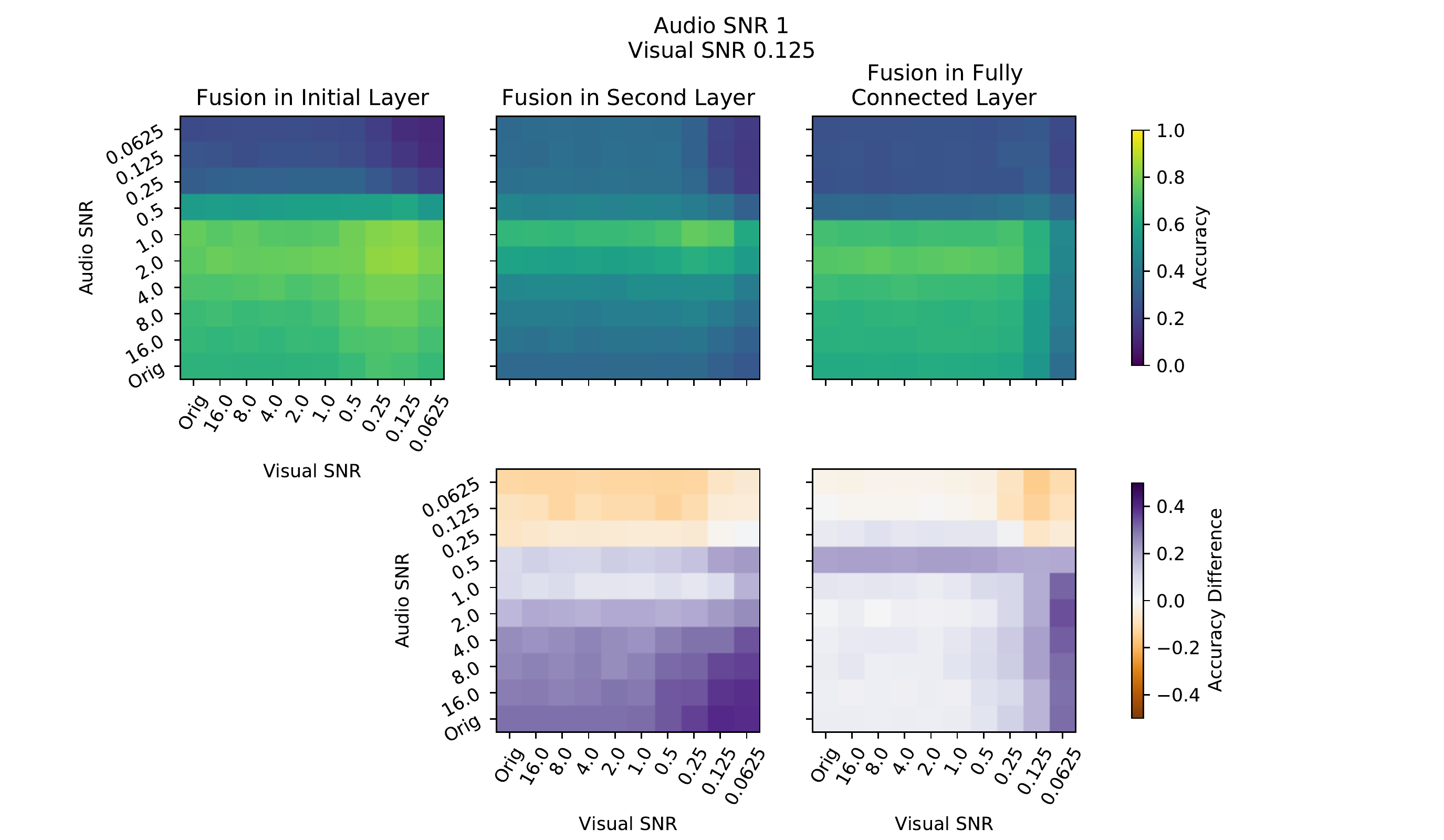}
  \vspace{-0.2in}
  \caption{Late fusion model where audio SNR = 1, and visual SNR = 0.125. Fusion in the initial layer outperforms the fusion in the fully connected layer for audio SNR values greater than 1.}
  \label{fig:late_fusion_a1_v0.125}
\end{figure}

\pagebreak

\subsection{Final Layer Activation}
\label{apend:activation}

Additional examples of final layer activations.

\begin{figure}[H]
  \centering
  \includegraphics[width=1\textwidth]{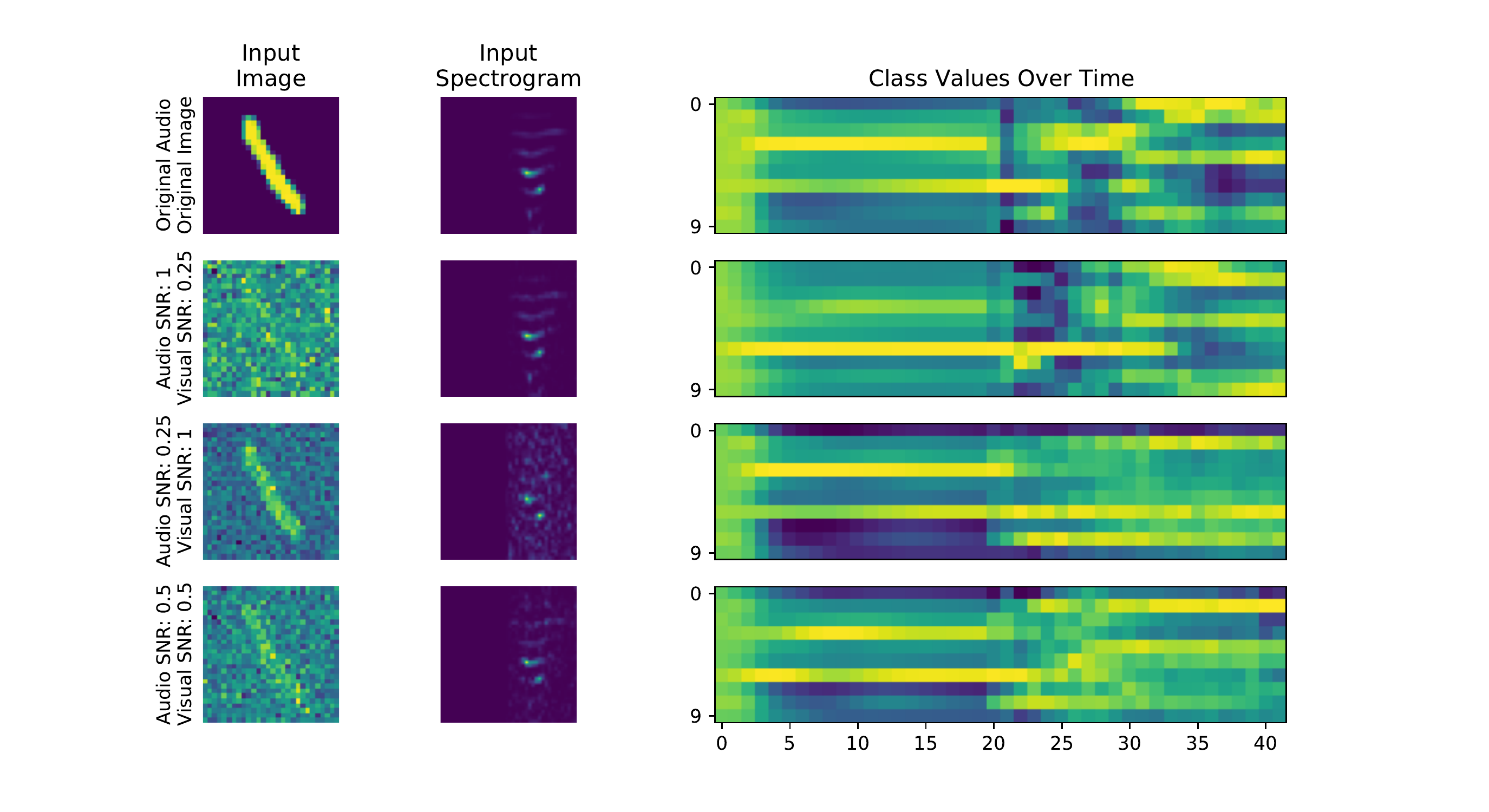}
  \vspace{-0.4in}
  \caption{ Additional example of values of the final layer of the multimodal layer across timesteps of the C-LSTM for a representative example at various signal to noise ratios.}
  \label{fig:class1}
\end{figure}

\begin{figure}[h!]
  \centering
  \includegraphics[width=1\textwidth]{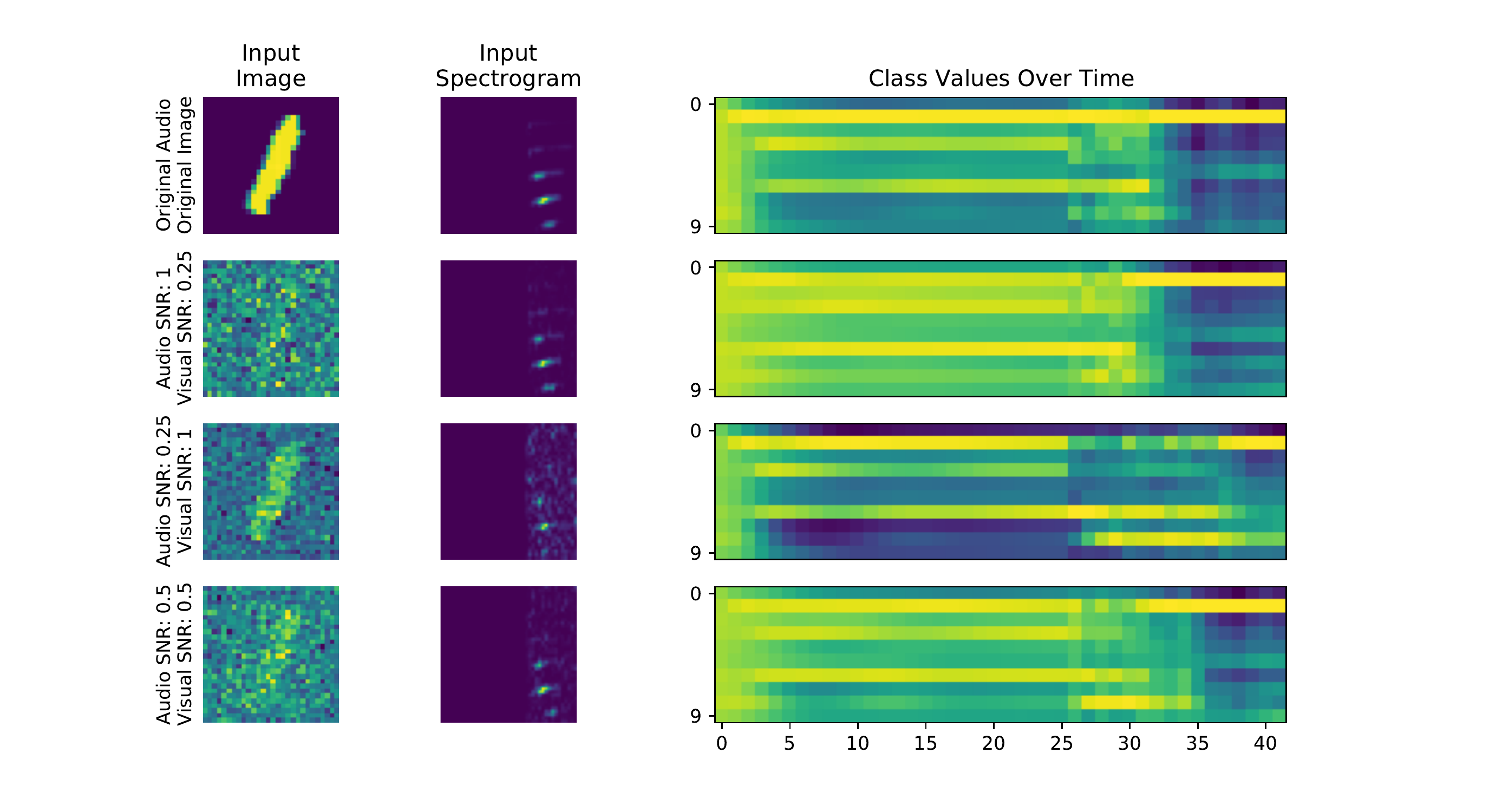}
  \vspace{-0.4in}
  \caption{ Additional example of values of the final layer of the multimodal layer across timesteps of the C-LSTM for a representative example at various signal to noise ratios.}
  \label{fig:class2}
\end{figure}

\begin{figure}[h!]
  \centering
  \includegraphics[width=1\textwidth]{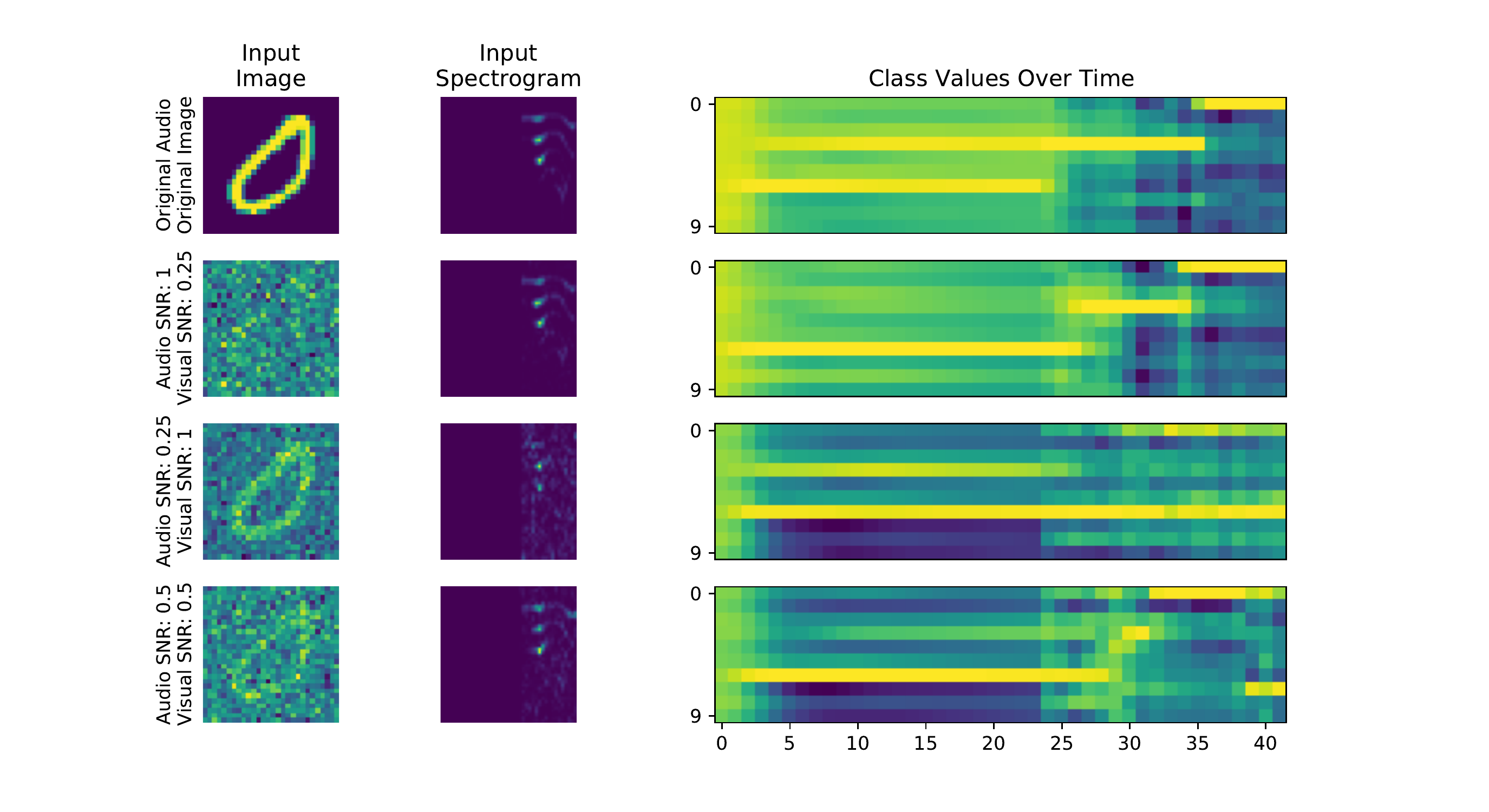}
  \vspace{-0.4in}
  \caption{ Additional example of values of the final layer of the multimodal layer across timesteps of the C-LSTM for a representative example at various signal to noise ratios.}
  \label{fig:class3}
\end{figure}

\begin{figure}[h!]
  \centering
  \includegraphics[width=1\textwidth]{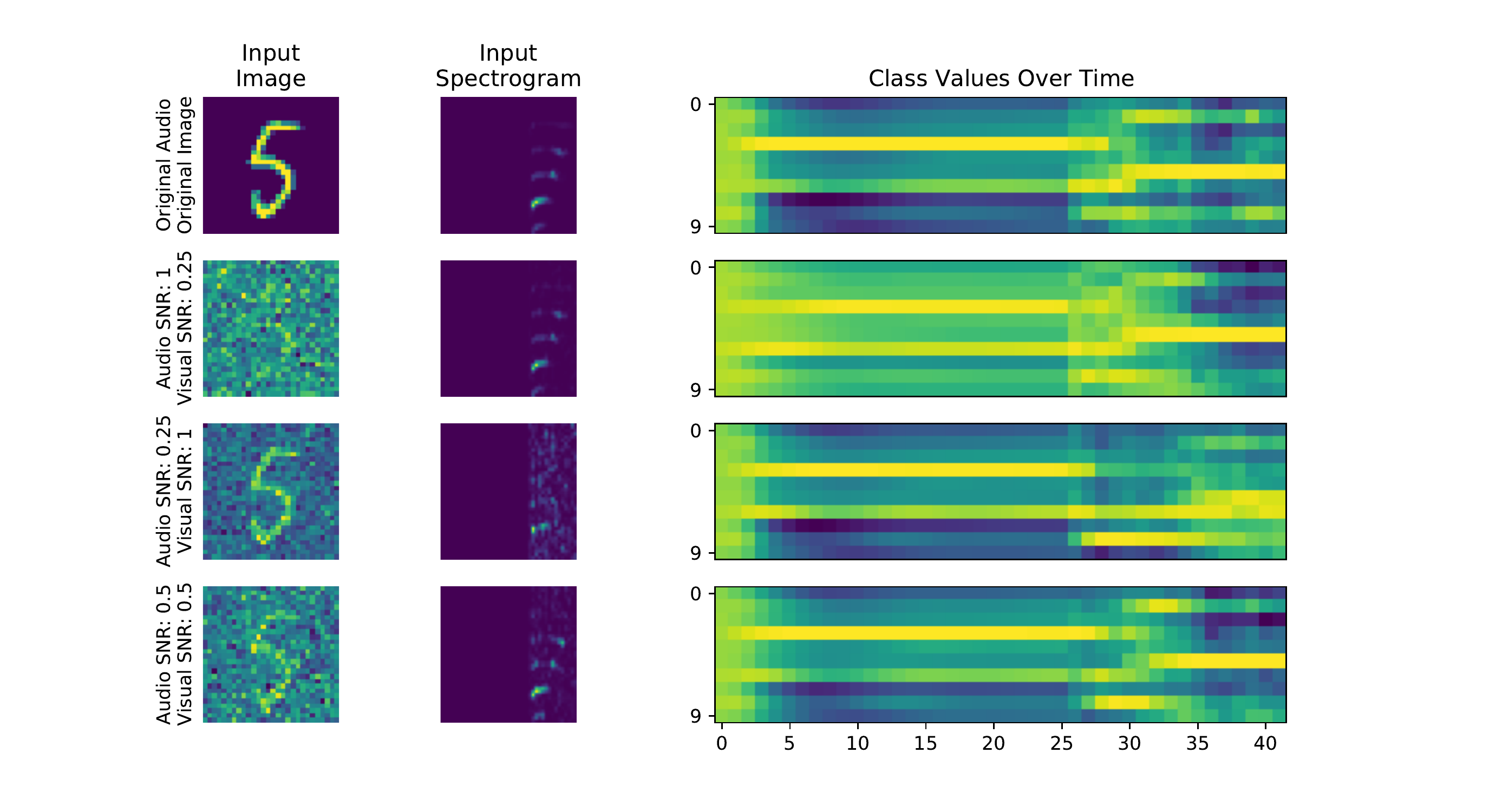}
  \vspace{-0.4in}
  \caption{ Additional example of values of the final layer of the multimodal layer across timesteps of the C-LSTM for a representative example at various signal to noise ratios.}
  \label{fig:class4}
\end{figure}

\begin{figure}[h!]
  \centering
  \includegraphics[width=1\textwidth]{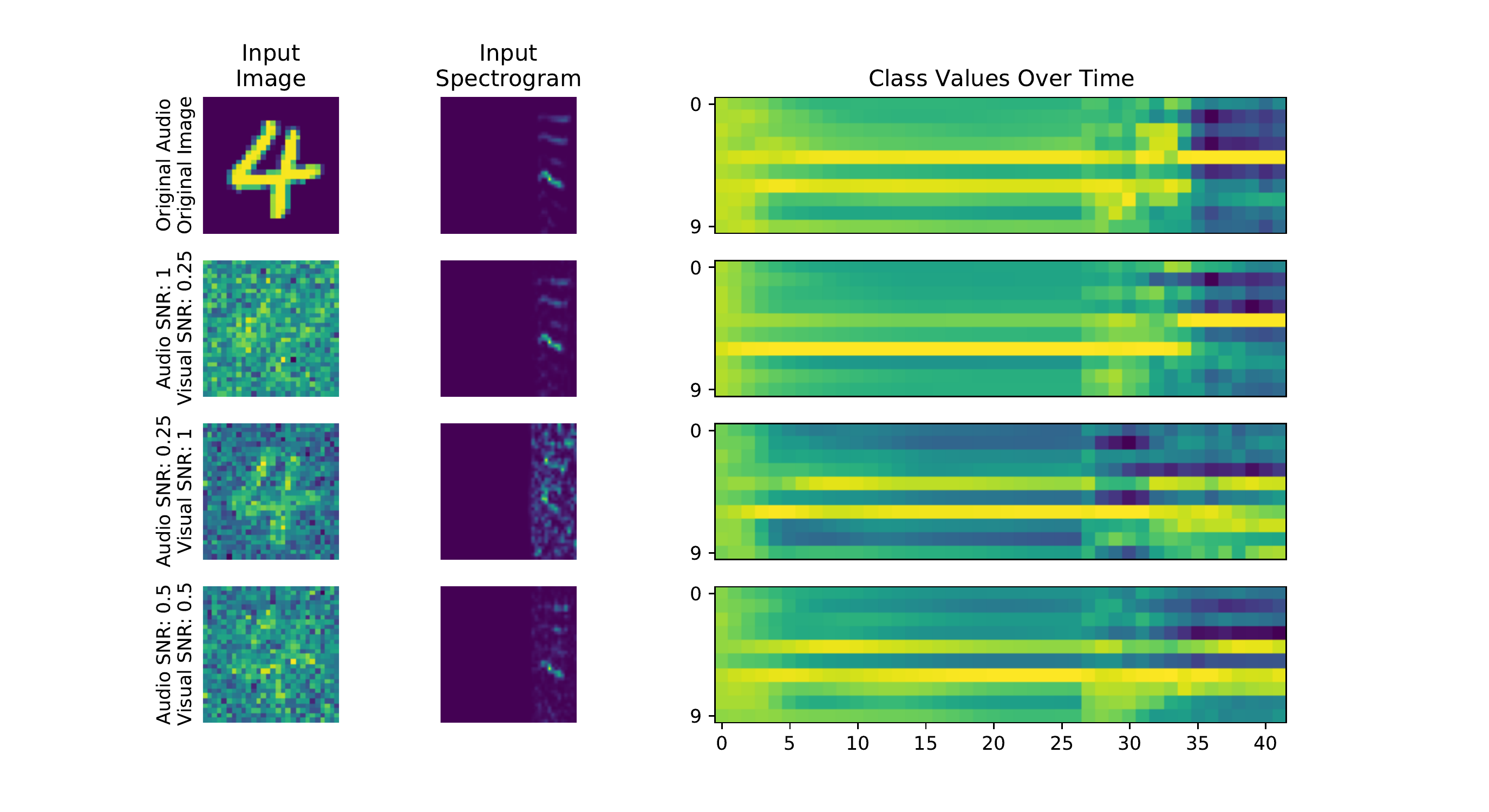}
  \vspace{-0.4in}
  \caption{ Additional example of values of the final layer of the multimodal layer across timesteps of the C-LSTM for a representative example at various signal to noise ratios.}
  \label{fig:class5}
\end{figure}

 \pdfoutput=1
\end{document}